\documentclass[letterpaper]{article} 
\usepackage{aaai25}  
\usepackage{times}  
\usepackage{helvet}  
\usepackage{courier}  
\usepackage[hyphens]{url}  
\usepackage{graphicx} 
\usepackage{natbib}  
\usepackage{caption} 
\usepackage{algorithm}
\usepackage{algorithmic}

\usepackage{multicol}
\usepackage{multirow}

\usepackage{amsmath}

\usepackage{newfloat}
\usepackage{listings}
\DeclareCaptionStyle{ruled}{labelfont=normalfont,labelsep=colon,strut=off} 
\floatstyle{ruled}
\newfloat{listing}{tb}{lst}{}
\floatname{listing}{Listing}

\newcommand{\sysname}{{MPruner}}

\title{MPruner: Optimizing Neural Network Size with CKA-Based Mutual Information Pruning}
\author{
	Seungbeom Hu\textsuperscript{\rm 1}\equalcontrib,
	ChanJun Park\textsuperscript{\rm 2}\equalcontrib,
	Andrew Ferraiuolo\textsuperscript{\rm 3}, 
	Sang-Ki Ko\textsuperscript{\rm 4}, 
	Jinwoo Kim\textsuperscript{\rm 5}, 
	Haein Song\textsuperscript{\rm 2}, 
	Jieung Kim\textsuperscript{\rm 6} 
}
\affiliations{
    \textsuperscript{\rm 1}Samsung SDS, South Korea\\
    seungbeom.hu@samsung.com\\
    \textsuperscript{\rm 2}Department of Computer Engineering, Inha University, Incheon, South Korea\\
    \{12181613, 22241401\}@inha.edu\\
    \textsuperscript{\rm 3}Independent Researcher, London, United Kingdom\\
    andrewferr@proton.me\\
    \textsuperscript{\rm 4}Department of Artificial Intelligence, University of Seoul, Seoul, South Korea\\
    sangkiko@uos.ac.kr\\
    \textsuperscript{\rm 5}Department of Computer Science and Engineering, Seoul National University, Seoul, South Korea\\
    jinwoo.kim@sf.snu.ac.kr\\
    \textsuperscript{\rm 6}Department of Computer Science, Yonsei University, Seoul, South Korea\\
    jieungkim@yonsei.ac.kr}

\begin{document}

\maketitle

\begin{abstract}
Determining the optimal size of a neural network is critical, as it directly impacts runtime performance and memory usage. Pruning is a well-established model compression technique that reduces the size of neural networks while mathematically guaranteeing accuracy preservation. However, many recent pruning methods overlook the global contributions of individual model components, making it difficult to ensure that a pruned model meets the desired dataset and performance requirements. To address these challenges, we developed a new pruning algorithm, \sysname, that leverages mutual information through vector similarity. \sysname\ utilizes layer clustering with the Centered Kernel Alignment (CKA) similarity metric, allowing us to incorporate global information from the neural network for more precise and efficient layer-wise pruning. We evaluated \sysname\ across various architectures and configurations, demonstrating its versatility and providing practical guidelines. \sysname\ achieved up to a 50\% reduction in parameters and memory usage for CNN and transformer-based models, with minimal to no loss in accuracy.

\end{abstract}

\section{Introduction}

Neural networks have revolutionized numerous fields, including computer vision and natural language processing~\cite{abiodun18, Elad23, otter20}, offering unprecedented levels of accuracy and functionality. However, with the success of neural networks also has come a significant increase in the size and complexity of neural networks, perhaps more so than strictly required: current neural networks often suffer from over-parametrization, where more neurons are used than necessary to effectively represent the dataset. Deployment of such over-parametrized models leads to unnecessary use of memory and power, which also translates to increased financial costs.

Over-parametrization is a significant issue across all model architectures, and the recent success of transformer-based models~\cite{Vaswani17, Lin22} has magnified this trend. These models have achieved high performance on various tasks~\cite{Dosovitskiy21, Khan22, Shamshad23} with extremely large networks, often exceeding billions of parameters. While transformers have opened a new chapter in performance, their size introduces new challenges, such as the need to fine-tune pre-trained models with data specific to new goals rather than training new models from scratch~\cite{Radiya-Dixit20, Tay22, Zaken22}. This trend forces users to maintain unnecessarily large pre-trained models~\cite{Fan24}, which limits the applicability of these powerful models, especially in resource-constrained environments like edge computing, where inference occurs on phones or other small devices. Consequently, there is a growing need for innovative solutions that try to find out the proper model size without sacrificing the benefits of advanced models.

One way to address this problem is through pruning~\cite{Cheng23, Han15, Molchanov16, Ma23, Frantar23, Sun24, Ganesh20, Fan21, Yang24, Liu24}, a well-known technique for compressing neural networks. While there is a substantial body of literature on this topic, most approaches focus on evaluating the importance of individual weights or optimizing an objective function subject to sparsity constraints. However, these methods often face two major challenges: (1) irregular memory access resulting from the selective pruning of weights within the same memory region, and (2) the lack of a multi-layer, cluster-wise approach with mathematical guarantees, due to the complexity and difficulty of analyzing the effects of layer clusters on model functionality. As a result, they offer a localized best-effort pruning solution, which is effective but fails to provide mathematical proof that the pruned size is optimal for the model and dataset.

\begin{figure}
\centering
\includegraphics[width=1\linewidth]{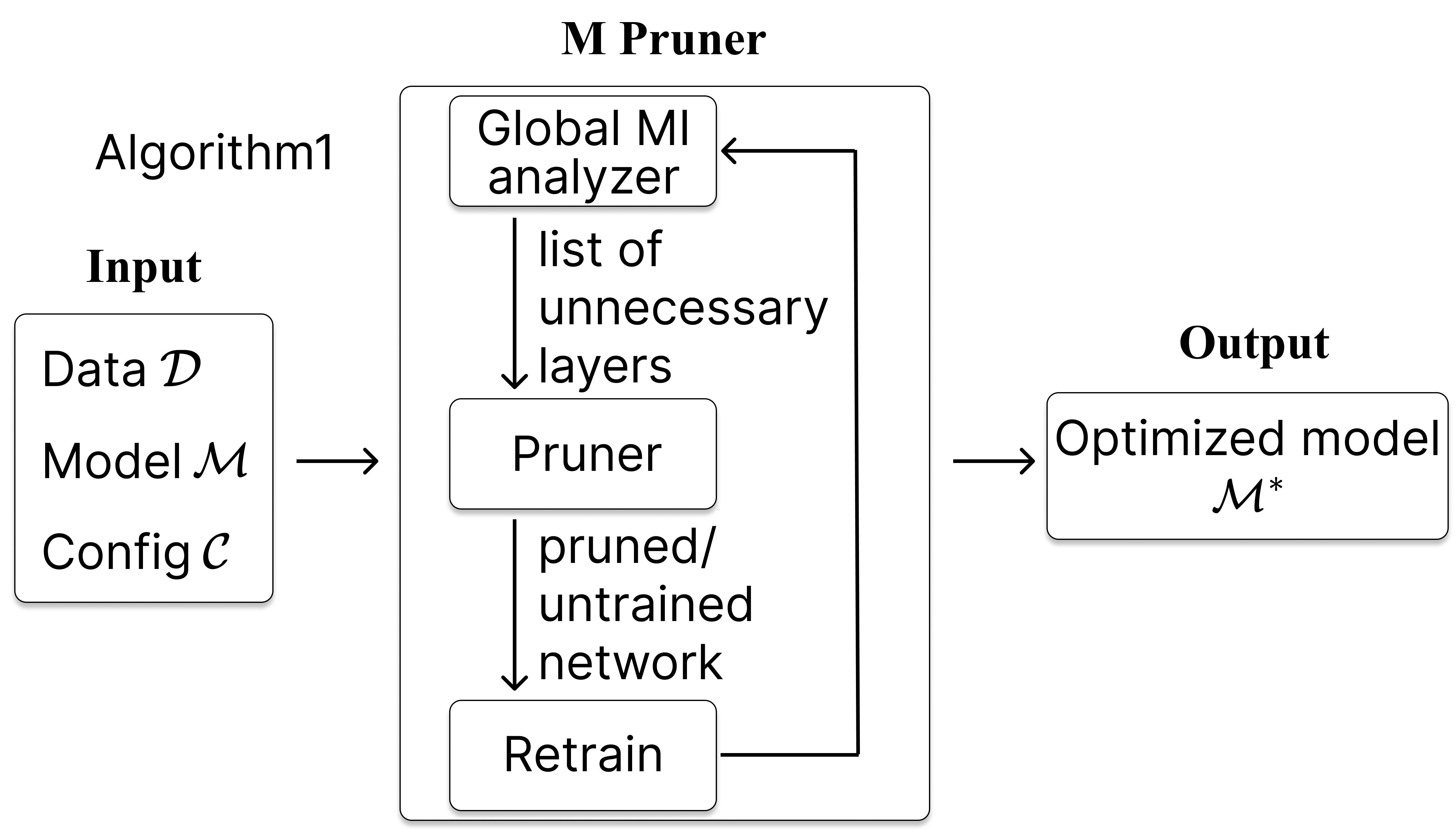}
\caption{Overview of \sysname}
\label{fig:system_overview}
\end{figure}

To overcome these limitations, we propose a novel pruning algorithm, \sysname, that leverages the Centered Kernel Alignment (CKA) similarity metric~\cite{Kornblith19} and global multi-layer collapsing, described in Figure~\ref{fig:system_overview}. \sysname\ first evaluates the contribution of each layer to the model, removes all unnecessary layers simultaneously, and then retrains the optimized model, resulting in a significantly smaller model size compared to the original. Unlike most previous pruning methods, \sysname\ uses an accuracy threshold rather than a sparsity level as the pruning criterion. \sysname\ also determines how many layers can be mathematically and safely pruned while staying within the specified threshold. We validated \sysname\ by applying it to various architectures, including CNN and transformer-based models, across a wide range of configurations. We also compared and discussed \sysname\ with earlier pruning methods, including mutual information-based techniques~\cite{Ganesh20, Fan21, Guo23, Yang24, Ma23}, and integrated \sysname\  with other state-of-the-art magnitude pruning methods for transformers~\cite{Sun24} to explore the potential of combining distinct approaches. Our experiments demonstrate that \sysname\ is an effective tool for pruning the network to an optimal size for the task with mathematical guarantees. Our experiments also show improvements in efficiency and performance, particularly in memory usage, supported by solid mathematical evidence that we effectively prune unnecessary components. We also provide valuable insights and guidelines for effective application. This highlights the potential of our approach to advance the development and deployment of neural networks, making large and complex models more accessible and sustainable by reducing the required computing resources.

Consequently, our work makes three key contributions, summarized as follows:
\begin{itemize}
\item We introduce a novel pruning algorithm, \sysname, which leverages multi-layer cluster-wise information to select pruning candidates. Our approach utilizes comprehensive global information from pre-trained networks, incorporates a retraining phase, and is adaptable to various architectures and pruning configurations.
\item We present experimental results demonstrating the effectiveness of our method across various configurations and models. Specifically, we explore four configurations: modularization threshold, accuracy drop threshold, pruning methods, and training methods. Our experiments also cover multiple architectures, including CNNs (ResNet50 and ResNet152), encoder-only transformer architecture (BERT), encoder-decoder transformer architecture (CodeT5), and utilize diverse datasets such as ImageNet, Dair AI, Yahoo! Questions and Answers, and SQuAD.
\item We examine the feasibility of integrating our methodology with different structural pruning methods for transformers~\cite{Sun24} to illustrate how our approach can enhance existing techniques.
\end{itemize}

\section{Background and Related Works}

 A large body of work addresses neural network compression, aiming to reduce model size and computational demands while maintaining performance. Among them, three well-known compression methods are quantization, pruning, and knowledge distillation. These methods can be combined rather than chosen as alternatives, as they have orthogonal effects toward the same goal of compression. These techniques are crucial for deploying neural networks in resource-constrained environments, such as mobile devices and edge computing platforms. Our paper focuses on proposing a new pruning method to improve memory and power efficiency.

\subsection{Pruning}
Pruning is broadly categorized into unstructured and structured approaches~\cite{Cheng23}, each with distinct characteristics and advantages. We focus on structured pruning~\cite{He24}, which is extensively studied both for CNNs~\cite{Li17, Luo17, He17, Liu17, Lin19, Huang18, Ganesh20} and Transformers~\cite{Michel19, Voita19, Kovaleva19, Lagunas21, Ma23, Frantar23, Sun24, Guo23, Kim24, Ashkboos24, Fan24, Men24, Yang24, An24, Fan21}. This approach simplifies the network's architecture by removing redundant or less critical structures, thereby enhancing overall efficiency. Common techniques in structured pruning include filter pruning, which targets entire filters in convolutional layers to achieve structured sparsity and improve computational efficiency. Neuron pruning, another structural method, involves removing entire neurons in fully connected layers, significantly reducing model size while preserving core functionality. While structured pruning offers the advantage of regular sparsity patterns that are more compatible with optimization on specialized hardware, it often requires careful adjustments to maintain model performance and can lead to changes in network architecture that impact training dynamics. 

Among these approaches, some utilize inter-layer information, such as mutual information within the network, to identify optimal pruning locations. For instance, Mutual Information-based Neuron Trimming (MINT)~\cite{Ganesh20} examines the relationships between filters in consecutive layers of CNN models, removing weights or filters to reduce layer size while considering their relative importance. In the context of transformer models, especially large language models (LLMs), Fan et al.~\cite{Fan21} determine sparsity levels based on inputs. However, their method focuses primarily on local mutual information by evaluating similarities between adjacent layers and does not consider the contributions of multiple layers simultaneously. LLM-Pruner~\cite{Ma23} employs structural pruning to selectively remove non-critical coupled structures based on gradient information. SliceGPT~\cite{Ashkboos24} replaces each weight matrix with a smaller matrix using Principal Component Analysis. LaCo~\cite{Yang24} gradually merges similar layers from deep to shallow, setting a threshold to avoid excessive merging of layers. As a result, these methods concentrate on identifying which layers are least or most important and should be pruned or removed, rather than selecting unnecessary layers by analyzing the global multi-layer cluster-wise information of the entire model. Consequently, while they may perform well using local evidence, they cannot demonstrate that the pruned model is optimally sized for the given task.

In this context, \sysname\ introduces a novel multi-architectural layer pruning approach that significantly enhances the use of analytical metrics by incorporating a state-of-the-art layer similarity metric to guide pruning, overcoming the limitations of previous and related methods. Unlike earlier approaches, \sysname\ offers two key advantages. First, our method offers both mathematical and intuitive criteria to ensure that pruning safely preserves accuracy. This is achieved by leveraging a well-known metric, Centered Kernel Alignment (CKA)~\cite{Kornblith19}, to calculate layer similarities. Second, it calculates the global contribution of multi-layer clusters rather than focusing solely on pairs of adjacent layers. This allows our method to capture global information more effectively, ensuring a more comprehensive and accurate pruning process.

\paragraph{Centered Kernel Alignment (CKA) similarity score}

Centered Kernel Alignment (CKA)~\cite{Kornblith19} is a metric used to assess the similarity between two sets of representations. Equation~\ref{eq:cka} presents the formula for calculating the CKA value between the outputs of adjacent layers ($i$ and $j$). It quantitatively measures how well the activations of one layer align with those of another layer or how similar two sets of activations are. Previous research~\cite{Davari23} has demonstrated that CKA is effective for evaluating the similarity of different layers in neural networks, particularly due to its ability to capture complex and non-linear relationships between activation patterns compared to other methods. As a result, CKA is considered a powerful tool for analyzing and comparing neural network layers.

The process of calculating the CKA score, depicted in Equations~\ref{eq:kernel}-\ref{eq:cka}, begins with the outputs of two different layers in the network. For example, if $x_i$ and $x_j$ are the outputs of two adjacent layers, Gram matrices $K$ and $L$ are first computed using Equation~\ref{eq:kernel}. These matrices are then centered using a centering matrix (Equations~\ref{eq:centering} and~\ref{eq:centering_kernel}). Next, the Hilbert-Schmidt Independence Criterion (HSIC)~\cite{Chwialkowski14} is calculated using Equation~\ref{eq:hsic}, which evaluates the independence between two random variables. Finally, the CKA score between the two layers is obtained using Equation~\ref{eq:cka}.  The CKA score offers practical benefits in terms of applicability, as it only requires the output vectors of the networks, eliminating the need to look into the network's detailed structure, such as weight values. In this sense, it can be applied to a wide range of architectures.

\begin{equation}\label{eq:kernel}
    K = x_i x_i^T \quad L = x_j \ x_j^T
\end{equation}

\begin{equation}\label{eq:centering}
    H = I_n - \frac{1}{n}J_n\quad\text{(Centering\ Matrix)}
\end{equation}

\begin{equation}\label{eq:centering_kernel}
    \hat{K} = HKH \quad \hat{L} = HLH
\end{equation}

\begin{equation}\label{eq:hsic}
    HSIC(K,L)= \frac{1}{(n-1)^2}\ tr(\hat{K}H\hat{L}H)
\end{equation}
\begin{equation}\label{eq:cka}
 CKA(x_i,x_j) = \frac{HSIC(K,L)}{\sqrt{HSIC(K,K)\ HSIC(L,L)}} 
\end{equation}

Recent work, DeepArc~\cite{Ren23}, utilizes the CKA metric to demonstrate the potential for clustering multiple adjacent layers into modules within a neural network. They compute the centered Gram matrix for all inputs and then use CKA values to identify layers with similar functions. They also apply DeepArc to ResNet-based models and explore various uses for modularization, including its potential for pruning. However, the work does not provide a concrete algorithm for managing different pruning-related configurations and lacks evidence of broader applicability, including the feasibility of applying their method to transformer models. Additionally, their method for calculating CKA values has room for improvement in both accuracy and scalability. They first store all layer-wise outputs for the given inputs, sum the output values, and then calculate the CKA score. However, this approach can lead to misleading information due to potential overflow when accumulated values are large and when there are significant differences in similarities across individual outputs. Therefore, we develop pruning algorithms, \sysname, that can accommodate multiple configurations, models, and architectures using the CKA value as a core metric, and report our new findings through extensive experimentation.

\section{\sysname: A Pruning Method for Determining Optimal Model Size}\label{sec:method}

\begin{figure}
\centering
\includegraphics[width=\linewidth]{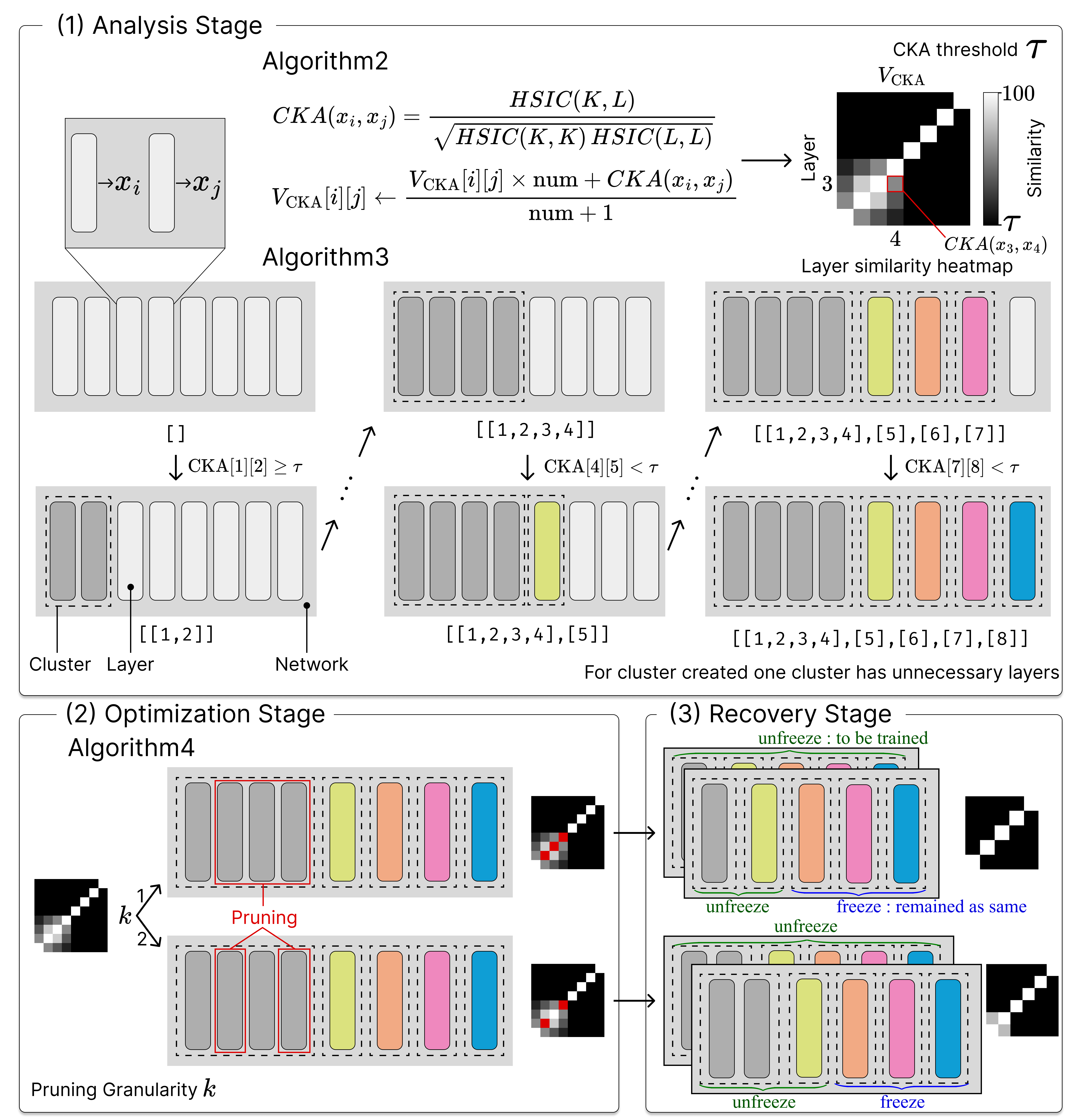}
\caption{Three Phases of \sysname: (1) Analysis Stage with CKA Similarity Score, (2) Optimization Stage with Multi-Layer Cluster Pruning, and (3) Recovery Stage with Retraining.}
\label{fig:process}
\end{figure}

\sysname\ consists of three phases, as depicted in Figure~\ref{fig:process}. The first phase is the \textbf{(1) analysis stage}, where the entire model is examined to provide multi-layer cluster-wise information. The second phase is the \textbf{(2) optimization stage}, where unnecessary and redundant layers with minimal impact on the model's functionality are removed. The third phase is the \textbf{(3) recovery stage}, in which the model's performance is restored through retraining.

The overall mechanism is illustrated in Algorithm~\ref{alg:cka_based_mi_pruning}. It takes five inputs: the model $\mathcal{M}$, the dataset $\mathcal{D}$, the hook position $\mathcal{L}$ to specify where to calculate the CKA-based mutual information metrics, the CKA threshold $\tau$ to determine the multi-layer clustering, the accuracy drop threshold $\gamma$ to determine when the algorithm should stop, and the freeze flag $b$ to indicate whether retraining should involve the entire network or focus only on the pruned parts. The algorithm sets the pruning granularity to 1 (line 4), meaning that all layers except one in redundant layer clusters will be removed. If set to 2, it prunes layers by skipping one after pruning one, effectively removing half of the redundant layers.

During each iteration, the algorithm first calculates and records the CKA values (line 6). If no layer clusters are found to perform similar functionalities in the model, the algorithm terminates (line 7). Otherwise, it invokes the pruner to remove redundant layers inside the multi-layer cluster according to the specified pruning granularity (line 8). The pruner then returns the pruned and untrained network $\mathcal{M}_{p, u}$ and the layers $ls_{f}$ that should be frozen if retraining is required. After retraining the pruned network, the algorithm calculates the accuracy of the updated version (lines 10 and 11), then adjusts either the model or the pruning granularity based on the observed accuracy drop (line 12).

Instead of specifying a sparsity level to prune the model parameters to a certain extent, the algorithm terminates when no more layers can be pruned or when the pruning granularity becomes too coarse. These conditions indicate that the network either has no layers left to prune or is no longer suitable for further pruning.

\begin{algorithm}[t]
\fontsize{9pt}{11pt}\selectfont 
\caption[Short label]{\texttt{\sysname}}\label{alg:cka_based_mi_pruning}
\begin{algorithmic}[1]
\REQUIRE model $\mathcal{M}$, dataset $\mathcal{D}$, hook positions $\mathcal{L}$, CKA threshold $\tau$, accuracy drop threshold $\gamma$, freeze flag $b$
\ENSURE pruned and retrained model
\STATE $\mathcal{M}^{*} \leftarrow \mathcal{M}$
\STATE $\mathcal{C} \leftarrow []$
\STATE $acc_o$, $acc_{pr}$ $\leftarrow$ \textbf{GetAccuracy}($\mathcal{M}, \mathcal{D}$)
\STATE $k \leftarrow 1$  (default pruning granularity)
\STATE \textbf{while} $k \le 3$ \textbf{do}
\STATE \ \ \ \ $\mathcal{C} \leftarrow$ \textbf{GetCandidates}($\mathcal{M}, \mathcal{L}, \tau, \mathcal{D}$)
\STATE \ \ \ \ \textbf{if} $\forall$ el ($\in \mathcal{C}$) . length(el) ==  1 \textbf{then} \textbf{break}
\STATE \ \ \ \ ($\mathcal{M}_{p, u}, ls_{f}$) $\leftarrow$ \textbf{Pruner}($\mathcal{M}, \mathcal{C}, k$) 
\STATE \ \ \ \ $\mathcal{M}_{f, u}$ $\leftarrow$ \textbf{Freezer}($\mathcal{M}_{p, u}, ls_{f}, \mathcal{L}, b$)
\STATE \ \ \ \ $\mathcal{M}^{*} \leftarrow $ \textbf{Trainer}($\mathcal{M}_{f, u}$)
\STATE \ \ \ \ $acc_{pr} \leftarrow$ \textbf{GetAccuracy}($\mathcal{M}^{*}, \mathcal{D}$)
\STATE \ \ \ \ \textbf{if} $acc_o - acc_{pr} \le \gamma$ \textbf{then} $\mathcal{M} \leftarrow \mathcal{M}^{*}$  \textbf{else} $k \leftarrow k + 1$ 
\STATE \textbf{end while}
\RETURN $\mathcal{M}$ 
\end{algorithmic}
\end{algorithm}

Two key functions in the algorithm are \textbf{GetCandidates} and \textbf{Pruner}. The \textbf{GetCandidates} function identifies layers within the network that perform similar functions. Algorithm~\ref{alg:generate_modules} outlines the essential steps for identifying these similarities, utilizing Algorithm~\ref{alg:calculate_cka_matrix}, which calculates the CKA values for all layers in the model and records them in the CKA matrix ($\mathcal{V}_{CKA}$). Once the CKA matrix is generated, pruning candidates are selected based on CKA value similarity: if the output vectors of two or more adjacent layers fall within a specified threshold, they are grouped together as a pruning candidate cluster ($ \mathcal{C}$).

\begin{algorithm}[t]
\fontsize{9pt}{11pt}\selectfont 
\caption[Short label]{\texttt{Calculate the CKA  matrix}}\label{alg:calculate_cka_matrix}
\begin{algorithmic}[1]
\REQUIRE $\mathcal{M}, \mathcal{L}, \mathcal{D}$
\ENSURE $\mathcal{V}_{CKA}$ 
\STATE $\mathcal{V}_{CKA}[][] \leftarrow \{0\}$
\STATE $\mathcal{D}_{seed}[] \leftarrow$ \textbf{GetSeed}($\mathcal{D}$)
\STATE \textbf{for} ($idx \in$ \textbf{len}($\mathcal{D}_{seed}$))  \textbf{do} 
\STATE \ \ \ \ $i \leftarrow \mathcal{L}[0]$
\STATE \ \ \ \ $output[] \gets  $\textbf{GetOutputs}($\mathcal{M}, \mathcal{L}, \mathcal{D}_{seed}[idx]$)
\STATE \ \ \ \ \textbf{for} $j \in \mathcal{L}[1:]$) \textbf{do}
\STATE \ \ \ \ \ \ \ \ $ cka \leftarrow $\textbf{GetCKA}($output[i], output[j]$)
\STATE \ \ \ \ \ \ \ \ $\mathcal{V}_{CKA}[i][j] \leftarrow (\mathcal{V}_{CKA}[i][j] * idx + cka) /  (idx + 1)$ 
\STATE \ \ \ \ \ \ \ \ $i \leftarrow j$
\STATE \ \ \ \ \textbf{end for}
\STATE  \textbf{end for}
\RETURN $\mathcal{V}_{CKA}$ 
\end{algorithmic}
\end{algorithm}

\begin{algorithm}[t]
\caption[Short label]{\texttt{Get Pruning Candidates}}\label{alg:generate_modules}
\begin{algorithmic}[1]
\REQUIRE $\mathcal{M}, \mathcal{L}, \tau, \mathcal{D}$
\ENSURE pruning candidate clusters $\mathcal{C}$
\STATE $\mathcal{V}_{CKA} \leftarrow$ \textbf{GetCKAMatrix}($\mathcal{M}, \mathcal{L}, \mathcal{D}$)
\STATE $\mathcal{C} \gets []$ 
\STATE $i \leftarrow \mathcal{L}[0]$
\STATE $cluster \gets \{i\}$ 
\STATE \textbf{for} $j \in \mathcal{L}[1:]$  \textbf{do} 
\STATE \ \ \ \ \textbf{if} {$\mathcal{V}_{CKA}[i][j] \geq \tau$} \textbf{then}  $cluster \cup \{j\} $ 
\STATE \ \ \ \ \textbf{else} \textbf{append} $cluster$ \textbf{to} $\mathcal{C}$ 
\STATE \ \ \ \ $cluster \gets \{j\}$ 
\STATE \ \ \ \ $i \gets j$ 
\STATE \textbf{end for} 
\STATE \textbf{return} $ \mathcal{C}$  
\end{algorithmic}
\end{algorithm}

The \textbf{Pruner}, as detailed in Algorithm~\ref{alg:pruning}, reduces the size of the target model based on the results from \textbf{GetCandidates} and the specified pruning granularity. If a layer cluster identified in the candidate set consists of two or more layers and the pruning granularity is set to $1$, our approach initially attempts to remove all layers in the cluster except the first one, based on the evidence that these layers perform similar roles within the model. For instance, if $n$ adjacent layers form a module ($p_i = {l_1,\ldots,l_n}$) and the output vectors from $l_1$ to $l_n$ fall within a similarity threshold $\tau$ or the given input values, merging these layers is expected to have minimal impact on the model's performance. 

However, we observe that excessive pruning sometimes leads to a noticeable drop in accuracy, particularly for CNN models. We believe this issue arises from having too many layers within a single cluster. To address this, we offer the option to adjust the pruning granularity to values other than $1$. If the granularity is set to $2$, the algorithm will attempt to remove every other layer in the cluster, effectively pruning the even-numbered layers. 

As defined in Algorithm~\ref{alg:cka_based_mi_pruning}, our algorithm adjusts the pruning ratio if the pruned model experiences a significant accuracy drop due to the reduction of multiple layers and fails to meet the accuracy threshold. We have found that such adjustments are generally unnecessary for transformer models. However, CNN models (such as ResNet~\cite{He16}) may require this step to mitigate accuracy loss.

\begin{algorithm}[t]
\caption[Short label]{\texttt{Pruning with Clustering Info.}}\label{alg:pruning}
\begin{algorithmic}[1]
\REQUIRE $\mathcal{M}, \mathcal{C}, k$
\ENSURE Pruned model $\mathcal{M}^{*}$, non-freeze layers $ls_f$ 
\STATE $ls_f\leftarrow \{\}$
\STATE $\mathcal{M}^{*}\leftarrow \mathcal{M}$
\STATE \textbf{for} $cluster \in \mathcal{C}[:-1]$   \textbf{do}
\STATE \ \ \ \ \textbf{for} $l_{idx} \in cluster[1:k][:-1]$ \textbf{do}
\STATE \ \ \ \ \ \ \ \ \textbf{if} dim({Out}$_{pre\_layer}$($l_{idx}$) == dim({In}$_{next\_layer}$($l_{idx}$) 
\STATE \ \ \ \ \ \ \ \ \textbf{then} $\mathcal{M}^{*} \leftarrow$ \textbf{DeleteLayer}($\mathcal{M}^{*}, l_{idx}$)
\STATE \ \ \ \ \ \ \ \ \textbf{else} $ls_f \leftarrow ls_f \cup \{l_{idx} - 1, l_{idx}\}$
\STATE \ \ \ \ \textbf{end for}
\STATE \textbf{end for}
\STATE \textbf{return} $\mathcal{M}^{*}$, $ls_f$
\end{algorithmic}
\end{algorithm}

\section{Experiments}

We conducted a range of evaluations by varying the inputs to \sysname. The primary goal of our experiments was not to argue that our method is superior to previous pruning techniques, but to confirm whether the multi-layer cluster pruning described in Section~\ref{sec:method} can achieve the optimal size, as supported by mathematical evidence measured by our CKA similarity score. All code and models were implemented in PyTorch. Due to the diversity of our experiments, we utilized multiple GPUs for analysis, optimization, recovery, and performance validation both before and after pruning. Specific details of the experimental setup are provided. As part of the inputs for \sysname, we set the CKA threshold ($\tau$) to 95\%, 98\%, and/or 99\%. The retraining process was conducted with a learning rate of 0.00005, a batch size of 32, and for 3 epochs. To ensure a thorough comparison, we report all intermediate results, not just the final input and output models of our top-level pruning algorithm~\ref{alg:cka_based_mi_pruning}.

Overall, we observed that our method effectively optimizes the size of both CNN and Transformer models with high scalability. The analysis stage operates in linear time, efficiently determining global multi-layer cluster-wise information regardless of the network's depth or width. Additionally, our approach demonstrated strong effectiveness for Transformer models, including both encoder-only and encoder-decoder architectures. We also found that our method can be seamlessly integrated with state-of-the-art magnitude pruning techniques~\cite{Sun24} without causing any accuracy loss in Transformer models.

\subsection{Transformer models}

\begin{figure}
\centering
\includegraphics[width=1\linewidth]{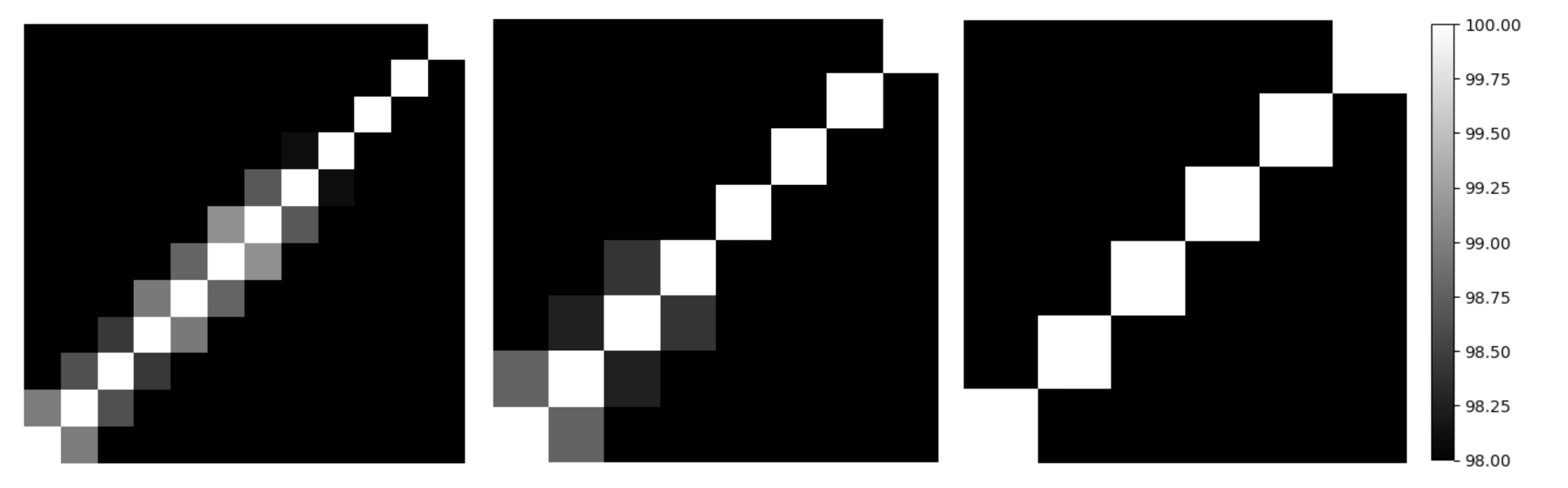}
(a) Dair AI Emotion
\includegraphics[width=1\linewidth]{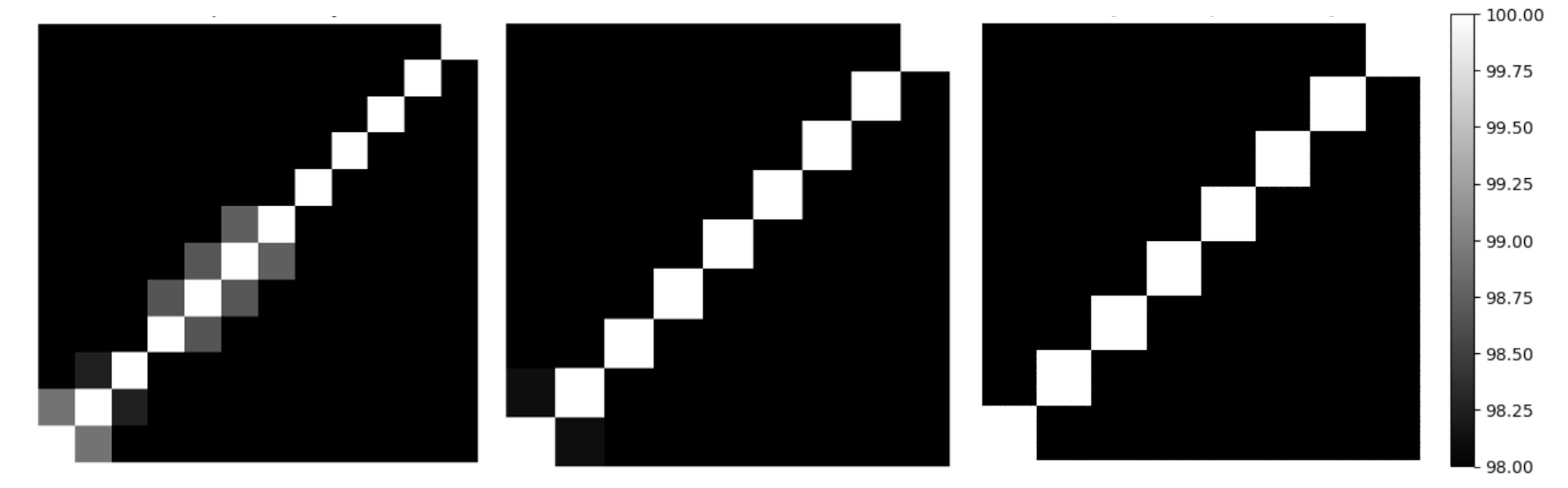}
(b) Yahoo Answers Topics
\caption{Clustering Heatmaps for Bert.}
\label{fig:bert-heatmap}
\end{figure}

\begin{table*}[!ht]
\centering
{\fontsize{8pt}{10pt}\selectfont 
\begin{tabular}{c|cccccc}
\hline
         & Method                           & Performance(\%)   & Evaluation time(s) & Training time(s) & Parameter(M) & Num. of Encoders \\
\hline 
\multirow{4}{*}{Dair AI Emotion} 
        & Baseline                          & 0.92                 &     14.8684        & 1579.8736       & 109.4869  & 12 \\
        & $\tau$ = 99\%, unfreeze, 1st iter.
                                            & 0.92        &  8.6455   &  1306.8720             & 95.3111  & 10\\
        & $\tau$ = 98\%, unfreeze, 1st iter.
                                            & 0.93 (+0.01)       & 7.0948   &  1062.2365            & 81.1354 & 8\\
        & $\tau$ = 98\%, unfreeze, 2nd iter.  
                                            & 0.93 (+0.01)       & 5.5214   &  796.2302             & 66.9596  & 6 \\
\hline
\multirow{3}{*}{Yahoo Answers Topics}  
        & Baseline                          & 0.70                 &     101.9891        & 41325.0308       & 109.4869  & 12 \\
        & $\tau$ = 99\%, unfreeze, 1st iter.
                                            & 0.70       &    101.9891   &  N.A.            & 109.4869 & 12\\
        & $\tau$ = 98\%, unfreeze, 1st iter.
                                            & 0.71 (+0.01)       & 79.4448   &  31593.5162            & 88.2263 & 8\\
\hline
\end{tabular}}
\caption{Bert Compression.}\label{tab:bert-result}
\end{table*}

\begin{figure}
\centering
  \begin{minipage}{.5\linewidth}
    \centering
\includegraphics[width=1\linewidth, height=2.7cm]{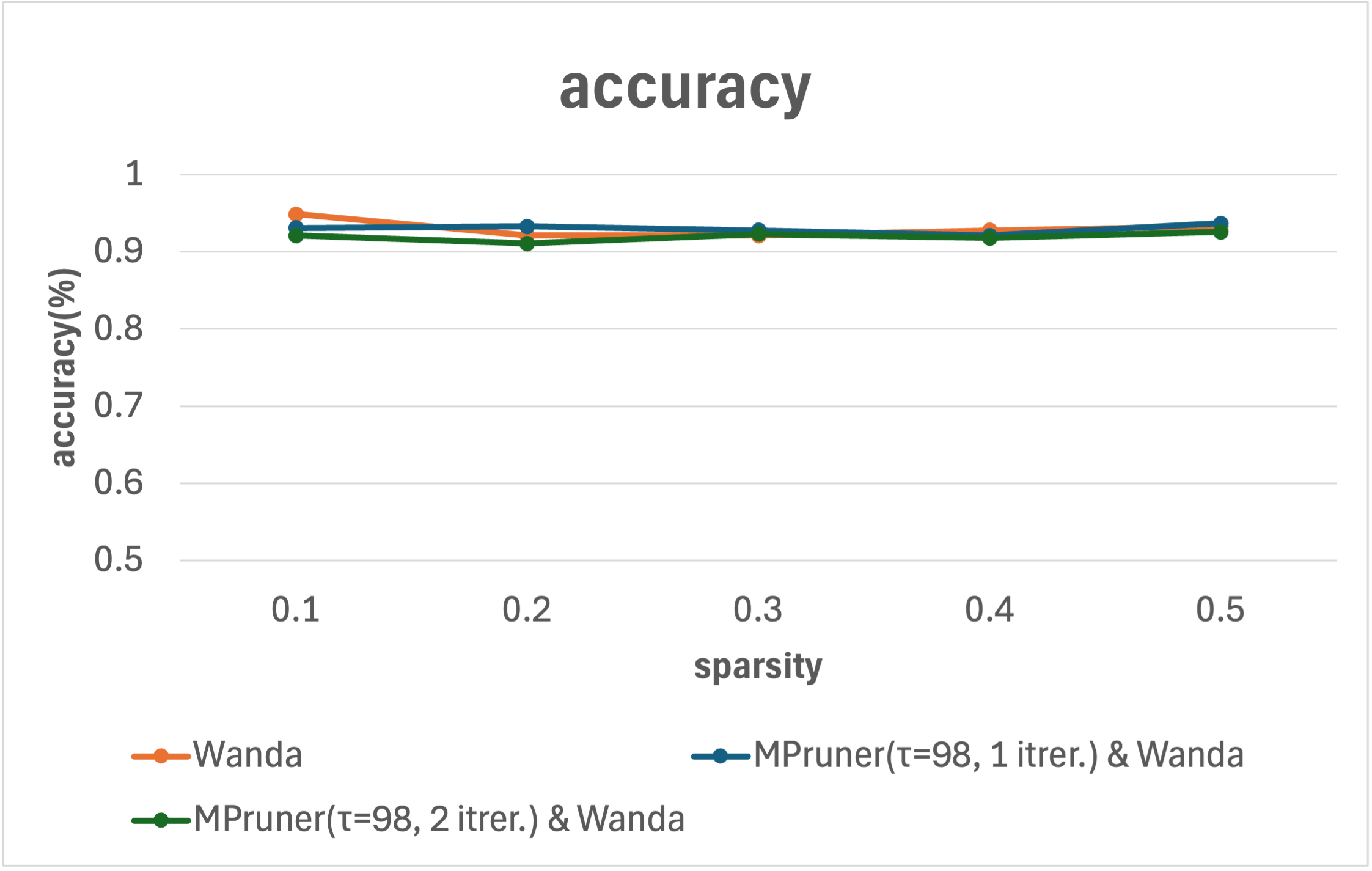}
  \end{minipage}%
  \begin{minipage}{.5\linewidth}
    \centering
  \includegraphics[width=1\linewidth, height=2.7cm]{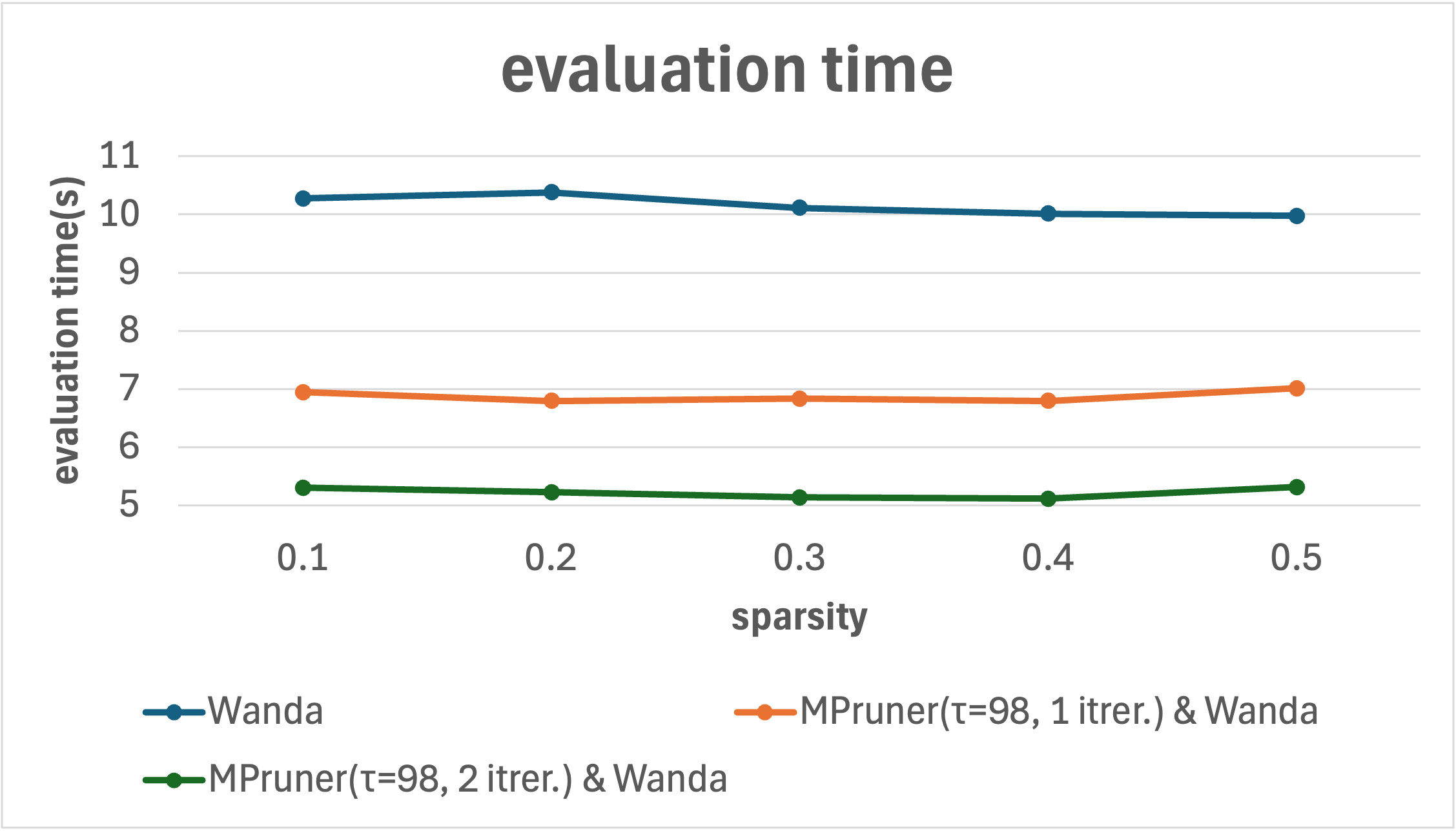}    
  \end{minipage}
\caption{Result of Wanda~\cite{Sun24}, and Integration of Wanda and \sysname.}
\label{fig:wanda-comparison}
\end{figure}

\paragraph{BERT-based} We performed two experiments with a BERT-based model. For small-scale data and label-related compression, we used a pretrained language emotion analysis model with six classes (Dair AI Emotion)\footnote{\url{https://huggingface.co/datasets/dair-ai/emotion}}. This dataset was divided into 16,000 training samples, 1,000 validation samples for CKA analysis, and 1,000 evaluation samples. For large-scale data and label-related experiments, we used a pretrained news article topic classification model with ten classes (Yahoo Answers Topics)\footnote{\url{https://huggingface.co/datasets/yahoo_answers_topics}}. This dataset was split into 1,400,000 training samples, 30,000 validation samples for CKA analysis, and 30,000 evaluation samples. The experiments were conducted on two different setups: the Dair AI model was tested on an NVIDIA GeForce RTX 3070 Ti 8GB GPU with a 13th Gen Intel(R) Core(TM) i7-13700 processor and 16GB of RAM, while the Yahoo Answers Topics model was tested on an NVIDIA GeForce RTX 4090 24GB GPU with a 14th Gen Intel(R) Core(TM) i9-14900 processor and 64GB of RAM.

Table~\ref{tab:bert-result} presents the experimental results. For a straightforward comparison, we disabled the freezing option in our configuration. We also evaluated several aspects affected by $\sysname$: performance (accuracy), evaluation time (the total inference time on our validation dataset), and training time (the duration required for training with a batch size of 32 and 3 epochs). Our method shows promising results, effectively reducing the model size by up to 50\% in terms of the number of encoders without compromising accuracy. Also, the final experimental results indicate that \sysname had almost no effect on accuracy. Meanwhile, inference speed, measured as the total inference time on the validation dataset, improved by 62.04\% and 22.10\%, and training time was reduced by 49.60\% and 23.55\%, respectively. Although the compression ratio varies depending on the dataset size, our methodology reliably identifies layers that can be collapsed, ensuring minimal loss of accuracy with mathematical rigor, as depicted in Figure~\ref{fig:bert-heatmap}. The figure displays the clustering results for both experiments: from left to right, showing BERT-base, the results after a single pruning iteration, and the final compression outcome. The proposed methodology in this paper involves iterative clustering and pruning until no further clustering of two or more encoders is possible. This heatmap illustrates the gradual reduction in layer similarity, providing evidence that $\sysname$ prunes the target network conservatively but correctly. The pruning is performed up to the point where the model remains safe and continues to offer nearly identical functionalities before and after pruning.

\paragraph{Potential for Integration with Other Tools} 
\sysname\ is designed to identify the optimal pruning point that maintains mathematical guarantees, rather than pushing the network to its limits. Consequently, it does not focus solely on minimizing the number of parameters, layers, or memory usage. To validate the safety and effectiveness of our pruning results, we conduct stress tests and apply an established pruning technique~\cite{Sun24} to both the original and pruned networks produced by \sysname. Our findings, illustrated in Figure~\ref{fig:wanda-comparison}, show that the accuracy of the pruned network is maintained at similar sparsity levels compared to the original model when pruned. Additionally, the reduced size of the network produced by \sysname\ enables its combination with Wanda, which further decreases inference time by eliminating additional weights compared to using Wanda alone. This empirically demonstrates that \sysname\ effectively removes unnecessary components while preserving network functionality and can enhance results when integrated with other pruning methods.

\paragraph{CodeT5}

\begin{table}[t]
\centering
{\fontsize{8pt}{10pt}\selectfont 
\begin{tabular}{c|ccc}
\hline
    & Baseline & 1st iter. & 2nd iter. \\ 
\hline
Num. of encoder	& 12 & 9 & 7 \\
Num. of decoder	& 12 & 5 & 3 \\
Num. of parameter	& 222M	& 135M	& 109M  \\
BLEU	& 0.2268	& 0.2120	& 0.2009 \\
ROUGE-1 F1	& 0.5177	& 0.5027	& 0.4929 \\
ROUGE-2 F1	& 0.3033	& 0.2880	& 0.2765 \\
ROUGE-L F1	& 0.4812	& 0.4697	& 0.4601 \\
ROUGE-Lsum F1	& 0.4812	& 0.4702	& 0.4601 \\
METEOR	& 0.4894	& 0.4702	& 0.4590 \\
\hline
\end{tabular}}
\caption{Compression on CodeT5}\label{tab:codet5-result}
\end{table}

\begin{table*}[t]
\centering
{\fontsize{8pt}{10pt}\selectfont 
\begin{tabular}{c|ccccc}
\hline
         & Method                           & Params./Mem. Pruned(\%)    & Performance(\%) & Num. Layers  \\ 
\hline 
\multirow{6}{1.5cm}{ResNet50\\[.1ex] ILSVRC2012} 
        & Baseline                          & N.A.                  &     76.75          & 16  \\
        & ThiNet~\cite{Luo17}               & 50.45/N.A.                 &     71.01 (-5.74)          & 16 \\
        & MINT~\cite{Ganesh20} ($\delta=0.1$) & 43.00/71.50                &     71.50  (-5.25)                  & 16 \\
        & Ours ($\tau$ = 98\%), unfreeze, 1st iter., k=1)
                                            & 5.4                   & 75.70 (-1.05)          & 14  \\
        & Ours ($\tau$ = 98\%), unfreeze, 2nd iter., k=1) 
                                            & 6.56                  & 74.75  (-2.00)       & 13 \\
\hline
\multirow{5}{1.5cm}{ResNet152\\[.1ex] ILSVRC2012}  
        & Baseline                          & N.A.                 &    76.82              & 50 \\
        & Ours ($\tau$ = 99\%), unfreeze, 1st iter., k=1)
                                            & 47.32                  &    73.37 (-3.45)       & 26 \\
        & Ours ($\tau$ = 98\%), unfreeze, 1st iter., k=1)
                                            & 54.74                 &    72.16 (-4.66)     & 22 \\
        & Ours ($\tau$ = 99\%), unfreeze, 1nd iter., k=2) 
                                            & 34.33                  &    74.96 (-1.86)        & 33 \\
        & Ours ($\tau$ = 98\%), unfreeze, 1nd iter., k=2)
                                            & 36.18                  &    74.93 (-1.89)        & 32 \\        
\hline
\end{tabular}}
\caption{CNN Models Compression.}\label{tab:cnn-result}
\end{table*}

\begin{figure}
\centering
\includegraphics[width=1\linewidth]{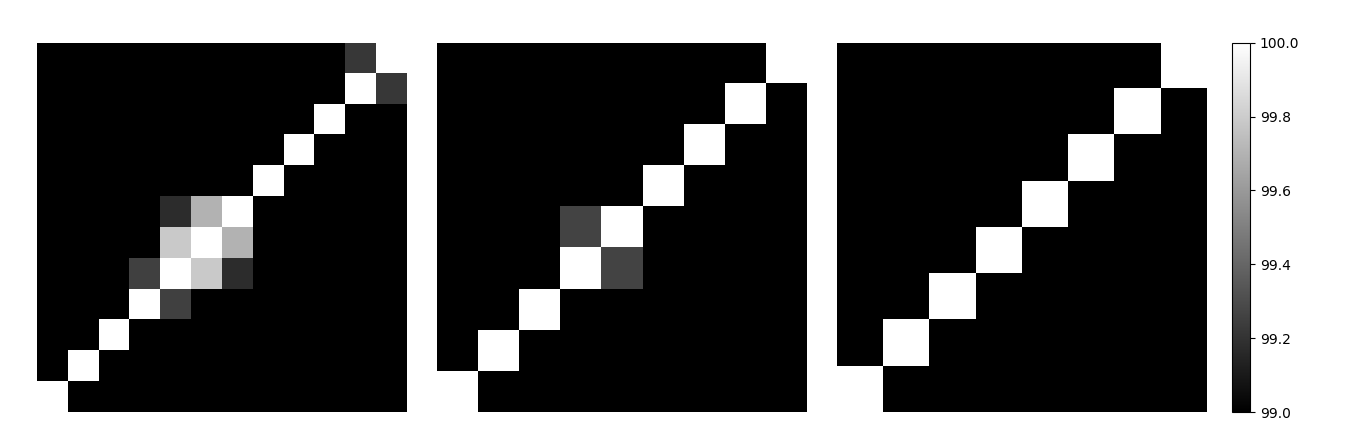}
(a) Encoder
\includegraphics[width=1\linewidth]{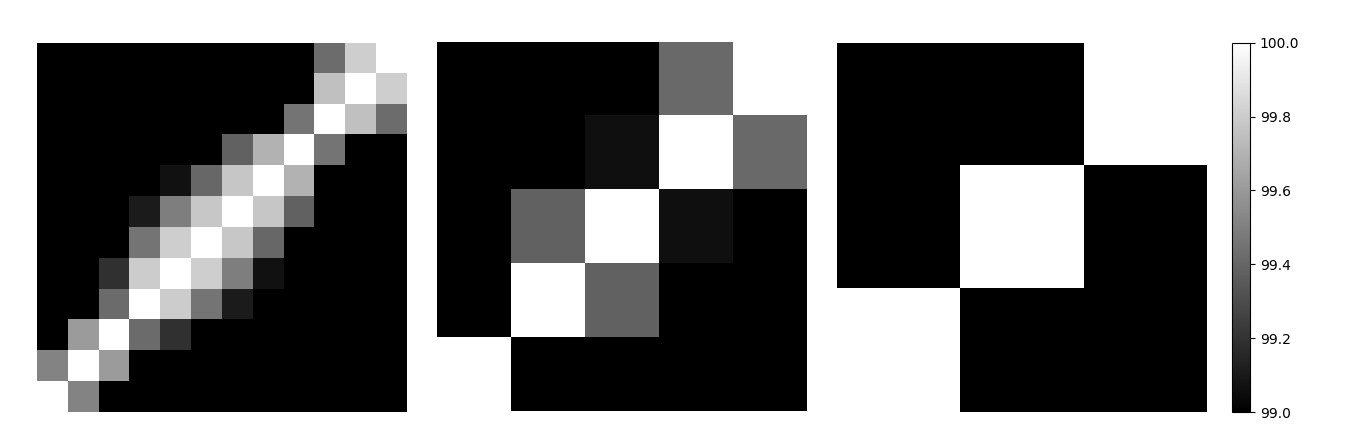}
(b) Decoder
\caption{Clustering Heatmaps for CodeT5.}
\label{fig:codet5-heatmap}
\end{figure}

Another experiment evaluates the applicability and impact of our method on a natural language generation model, specifically the T5-base model using the SQuAD dataset\footnote{\url{https://huggingface.co/datasets/rajpurkar/squad}}. The experiment was conducted on an NVIDIA RTX A6000 48GB GPU, an Intel(R) Xeon(R) W5-2455X processor, and 256 GB of RAM. Unlike previous benchmarks that used accuracy as the sole performance metric, we employed three well-known metrics—BLEU, ROUGE, and METEOR~\cite{Liu16}—to assess the effectiveness of our approach. We set the CKA similarity threshold $\tau$ to 99\% and the maximum allowable accuracy drop threshold $\gamma$ to 0.02\% for the BLEU score. 

With this threshold, we found that \sysname\ could effectively prune the network to the point where no clusters remained after two iterations. The clustering information from each iteration is illustrated in Figure~\ref{fig:codet5-heatmap}. The results, also detailed in Table~\ref{tab:codet5-result}, show a significant reduction in the size of the CodeT5 model, particularly in the decoder layers. \sysname\ identified that approximately two-thirds of the decoder layers were unnecessary, which the results confirm. Even for the encoder layers, \sysname\ demonstrated that nearly half were superfluous. Overall, the results suggest that CodeT5 is significantly larger than needed for the SQuAD dataset, and \sysname\ effectively reduces the model size with both mathematical justification and empirical evidence. 

We have not tested \sysname\ on large language models, such as LLaMa, but the pruning ratios achieved in our three experiments are comparable to or exceed those reported in previous works~\cite{Ma23, Liu24, Yang24, Fan24, Kim24, Men24}, despite the distinct nature of our methodology compared to these tools. Since \sysname\ determines the pruning ratio based on the model's size and dataset characteristics, we can expect that large language models may achieve higher pruning ratios depending on their specific tasks and requirements.

\subsection{CNN models}

We applied \sysname\ to two ResNet models, ResNet50 and ResNet152, using the ILSVRC-2012 dataset\footnote{\url{https://www.image-net.org/challenges/LSVRC/2012/}}. Both models were run on an NVIDIA RTX A4000 16GB GPU, an Intel(R) Xeon(R) W5-2455X processor, and 128 GB of RAM. The results are summarized in Table~\ref{tab:cnn-result}. Compared to experiments with transformer models, we observed that the accuracy drop occurs more rapidly with ResNet models. We set the accuracy drop threshold $\gamma$ to 2\%, but \sysname\ quickly reached this threshold within one iteration. Even with adjustments to the pruning granularity for ResNet152, we encountered a significant accuracy drop compared to what we observed with transformer models.

Comparing \sysname\ with recent previous structured pruning methods~\cite{Luo17, Ganesh20} on ResNet50, our approach effectively preserves accuracy even at the highest pruning granularity ($k=1$), though it can only safely prune a limited number of layers. Additionally, MINT faces scalability issues with ResNet152, as datasets larger than 5,000 samples exceed available memory and calculating mutual information for every layer pair can take several days. 

In summary, while CKA scores are more suitable for measuring redundancy in transformer models, our method still offers reasonable safety and scalability for pruning CNN models.

\subsection{TinyNas} 

We also evaluated the applicability of our approach to a network architecture search (NAS) algorithm, TinyNAS~\cite{Wang23a, Wang23b}, to determine if it could identify optimal model sizes. Our findings indicate that TinyNAS does not produce meaningful multi-layer clusters in its results. Considering TinyNAS aims to optimize architectures for resource-constrained devices, we assume it assesses layer similarities in the synthesized models with the given dataset. We plan to integrate our tool with additional NAS algorithms, such as Cream, to further explore and enhance its applicability.

\section{Conclusion and Future Works}

We propose a novel pruning method that collapses layers while regulating the pruning ratio according to specified accuracy thresholds, using multi-layer cluster-wise information to optimize neural network size. This method leverages layer similarities in architectures such as CNNs and transformers, removing unnecessary layers based on CKA similarity values between layer clusters. Our experiments show that this approach is highly effective for transformer models and efficiently reduces the size of large CNN models by eliminating unnecessary components. Additionally, our results suggest that our method can be seamlessly integrated with other pruning techniques to further reduce both model size and inference time.

In the future, we plan to explore alternative metrics beyond CKA for identifying layers to prune, aiming to improve compatibility with various architectures. We also intend to integrate our approach with advanced NAS tools and conduct further evaluations on large-scale language models, such as LLaMa and GPT, to assess the scalability and broader applicability of our method.
\bibliography{refs}

\appendix
\newpage

\section{Supplementary Materials}

The following sections present additional experiments involving \sysname\ that are not covered in the main text.

\section{ResNet}

\begin{itemize}
    \item Hareware setting:
    \begin{itemize}
        \item CPU : Intel(R) Xeon(R) W5-2455X processor
        \item GPU : NVIDIA RTX A4000 16GB GPU
        \item MEMORY : 128 GB
    \end{itemize}
    \item Dataset : ILSVRC-2012
    \item Model :
    \begin{itemize}
        \item microsoft/resnet152
        \item microsorf/restnet50
    \end{itemize}
\end{itemize}

\subsection{ResNet152}

This experiment compares a model that was fine-tuned using Microsoft’s ResNet152 for 3 epochs with a model that is compressed by \sysname. In this experiment, we set CKA thresholds as 98 and 99, followed by 3 epochs of retraining and unfreezing/freezing. The batch size was set to 64, the optimizer used was Adam, and the learning rate was 0.0002. The layer pruning process of the model was based on the similarity of the intermediate layers, specifically focusing on the ResNetBottleneck layer. The results of the experiment are in Table~\ref{suptab:resnet152}.

\begin{table*}[t]
{\centering
{\fontsize{8pt}{10pt}\selectfont 
    \begin{tabular}{|l|l|l|l|l|l|l|l|l|}
    \hline
        Stage & CKA Threshold & 1epoch & 2epoch & 3epoch & Num. of layers & Parameter & Eval time(s) & Freeze \\ \hline
        Original & N/A & N/A & N/A & 76.824 & 50 & 60.192808 M & 259.9679s & N/A \\ \hline
        \multirow{4}{1.5cm}{All\\[.1ex]1st iteration\\[.1ex]k=1} & $\tau=$99 & 72.094 & 72.926 & 73.374 (-3.45\%) & 26 & 31.709224 M & 180.18139s & 0 \\ 
        ~ & $\tau=$98 & 70.318 & 72.0 & 72.164 (-4.66\%) & 22 & 27.240488 M & 169.60528s & 0 \\ 
        ~ & $\tau=$99(freeze) & 72.604 & 73.494 (-3.33\%) & 73.176 (-4.33\%) & 26 & 31.709224 M & 180.17326s & 16 \\ 
        ~ & $\tau=$98(freeze) & 70.826 & 71.894 & 72.294 (-4.53\%) & 22 & 27.240488 M & 169.92417s & 15 \\ \hline
        \multirow{4}{1.5cm}{Half (Odd)\\[.1ex]1st iteration\\[.1ex]k=2} & $\tau=$99 & 74.758 & 74.960 (-1.864\%) & 74.622 & 33 & 39.529512 M & 202.43821s & 0 \\ 
        ~ & $\tau=$98 & 74.768 & 74.926 (-1.898\%) & 74.652 & 32 & 38.412328 M & 201.78150s & 0 \\ 
        ~ & $\tau=$99(freeze) & 74.572 & 74.956 (-1.868\%) & 74.818 & 33 & 39.529512 M & 202.33958s & 9 \\
        ~ & $\tau=$98(freeze) & 74.51 & 74.688 (-2.136\%) & 74.6859 & 32 & 38.412328 M & 201.15831s & 8 \\ \hline
        \multirow{4}{1.5cm}{All\\[.1ex]2nd iteration\\[.1ex]k=1}  
        & $\tau=$99 & 70.226 & 70.39 & 70.53 & 20 & 25.84324 M & 162.24440s & 0 \\ 
        ~ & $\tau=$98 & 67.168 & 68.49 & 68.822 & 14 & 19.977256 M & 139.80341s & 0 \\ 
        ~ & $\tau=$99(freeze) & 68.984 & 69.512 & 70.054 & 17 & 22.491688 M & 151.83072s & 9 \\ 
        ~ & $\tau=$98(freeze) & 70.35 & 70.61 & 70.254 & 17 & 23.328808 M & 149.61890s & 9 \\ \hline
        \multirow{4}{1.5cm}{Half (Odd)\\[.1ex]2nd iteration\\[.1ex]k=2}
        & $\tau=$99 & 73.02 & 73.38 & 73.106 & 26 & 33.663528 M & 181.37359s & 0 \\ 
        ~ & $\tau=$98 & 72.44 & 72.789 & 72.807 & 23 & 30.031912 M & 166.74806s & 0 \\ 
        ~ & $\tau=$99(freeze) & 73.672 & 73.842 & 73.64 & 27 & 33.663528 M & 181.26035s & 15 \\ 
        ~ & $\tau=$98(freeze) & 72.326 & 72.460 & 72.64 & 22 & 28.077608 M & 168.01924s & 8 \\ \hline
    \end{tabular}}
\caption{ResNet152 - ILSVRC-2012}\label{suptab:resnet152}}
\end{table*}

\subsection{ResNet50}

This experiment compares a model fine-tuned using Microsoft’s ResNet50 for 3 epochs with a model that is compressed by \sysname. In this experiment, we set CKA thresholds as 95, 98, and 99, followed by 3 epochs of retraining and unfreezing/freezing. However, the threshold of 99 was excluded as no pruning occurred. The batch size was set to 64, the optimizer used was Adam, and the learning rate was 0.0002. The layer pruning process of the model was based on the similarity of the intermediate layers, specifically focusing on the ResNetBottleneck layer. The results of the experiment are in Table~\ref{suptab:resnet50}.

\begin{table*}[t]
{\centering
{\fontsize{8pt}{10pt}\selectfont 
    \begin{tabular}{|l|l|l|l|l|l|l|l|l|}
    \hline
        Stage & CKA Threshold & 1epoch & 2epoch & 3epoch & Num. of layers & Parameter & Eval time(s) & Freeze \\ \hline
        \multirow{3}{1.5cm}{All\\[.1ex]1st iteration\\[.1ex]k=1}  & ~ & 77.2 & 77.082 & 76.754 & 16 & 25.557032 M & 147.0041 & ~ \\ \hline
        ~ & $\tau$=99 & (=original) & ~ & ~ & ~ & ~ & ~ & ~ \\ 
        ~ & $\tau$=98 & 75.698(-1.056\%) & 75.44 & 75.2719 & 14 & 24.159784 M & 136.8380 & 0 \\ 
        ~ & $\tau$=95 & 72.202 & 72.452(-4.302\%) & 72.432 & 10 & 21.574952 M & 100.2376 & 0 \\ \hline
        \multirow{3}{1.5cm}{All\\[.1ex]2nd iteration\\[.1ex]k=1}  & $\tau$=98 & 74.75 & 74.652 & 74.468 & 13 & 23.87972 M & 130.574 & 0 \\ 
        ~ & $\tau$=95 & 70.414 & 70.69 & 70.524 & 8 & 21.224488 M & 110.483 & 0 \\ 
        ~ & $\tau$=98(freeze) & 72.48 & 72.568 & 72.658 & 11 & 21.294888 M & 127.497 & 9 \\ 
        ~ & $\tau$=95(freeze) & 71.7539 & 71.364 & 71.566 & 9 & 21.645352 M & 116.2409 & 5 \\ \hline
    \end{tabular}}
\caption{ResNet50 - ILSVRC-2012}\label{suptab:resnet50}}
\end{table*}

\section{Dair AI Emotion}

\begin{itemize}
    \item Hareware setting:
    \begin{itemize}
        \item CPU : 13th Gen Intel(R) Core(TM) i7-13700
        \item GPU : NVIDIA GeForce RTX 3070 TI 8GB
        \item MEMORY : 16GB
    \end{itemize}
    \item Dataset : Dair AI Emotion
    \item Model : Bert-base / text classification
\end{itemize}

This experiment compares a model fine-tuned using BERT-base for text classification with 1 epoch to a model compressed by \sysname. Additionally, we conduct experiments with Wanda, comparing the results of applying Wanda alone versus using the integrated tool combining Wanda and \sysname. The sparsity levels for Wanda are 10\%, 20\%, 30\%, 40\%, and 50\%. In this experiment, CKA thresholds of 98 and 99 were used, followed by 3 epochs of retraining and unfreezing/freezing. The batch size was set to 32, the optimizer used was AdamW, and the learning rate was 5e-05. The layer pruning process was based on the similarity of BERT layers, specifically focusing on the encoder. The results of the experiment are as follows. 

The results from the basic experiments conducted solely with \sysname\ are shown in Figure~\ref{suptab:dairai}. Comparison with Wanda is described in Figure~\ref{suptab:dairaiwanda}.

\begin{table*}[t]
{\centering
{\fontsize{8pt}{10pt}\selectfont 
    \begin{tabular}{|l|l|l|l|l|l|l|l|l|}
    \hline
        Stage & CKA Threshold & Num. of enc. & Eval. loss & Accuracy & Eval. time(s) & Training time(s) & Parameter & Freeze \\ \hline
        \multirow{4}{1.5cm}{1st iteration}  & ~ & 12 & 0.3312 & 0.92 & 14.8684 & 1579.8736 & 109.486854 M & ~ \\ 
        ~ & $\tau$=99 & 10 & 0.4160 & 0.921(+0.001\%) & 8.6455 & 1306.872 & 95.31111 M & 0 \\ 
        ~ & $\tau$=98 & 8 & 0.3665 & 0.934(+0.014\%) & 7.0948 & 1062.2365 & 81.135366 M & 0 \\
        ~ & $\tau$=99(freeze) & 10 & 0.3125 & 0.926(+0.006\%) & 8.6366 & 1126.2001 & 95.31111 M & 6 \\ 
        ~ & $\tau$=98(freeze) & 8 & 0.3341 & 0.925(+0.005\%) & 6.9947 & 929.7921 & 81.135366 M & 4 \\ \hline
        \multirow{2}{1.5cm}{2nd iteration}  & $\tau$=98 & 6 & 0.4054 & 0.928(+0.008\%) & 5.5214 & 796.2302 & 66.959622 M & 0 \\ 
        ~ & $\tau$=98(freeze) & 6 & 0.4139 & 0.915(-0.005\%) & 5.3583 & 706.6606 & 66.959622 M & 3 \\ \hline
    \end{tabular}}
\caption{Bert - Dair AI Emotion, \sysname\ Only}\label{suptab:dairai}}
\end{table*}

\begin{table*}[t]
{\centering
{\fontsize{8pt}{10pt}\selectfont 
    \begin{tabular}{|l|l|l|l|l|l|l|}
    \hline
        ~ & CKA threshold & Sparsity level & Eval. loss & Accuracy (w/o training) & Evaluation time(s) & Parameter \\ \hline
        \multirow{5}{1.5cm}{Wanda} & N/A & 0.1 & 0.23041 & 0.949(+0.029\%) & 10.272 & 109.486854M \\ 
        ~ & N/A & 0.2 & 0.28538 & 0.921 & 10.382 & 109.486854M \\ 
        ~ & N/A & 0.3 & 0.308224 & 0.921 & 10.116 & 109.486854M \\ 
        ~ & N/A & 0.4 & 0.26523 & 0.928 & 10.0158 & 109.486854M \\ 
        ~ & N/A & 0.5 & 0.2619 & 0.933 & 9.9735 & 109.486854M \\ \hline
        \multirow{20}{1.5cm}{\sysname,\\[.1ex]Wanda} & $\tau$=98[1st iter.] & \multirow{4}{1.5cm}{0.1} & 0.38772 & 0.931 & 6.9494 & 81.135366M \\ 
        ~ & $\tau$=98(freeze)[1st iter.] & ~ & 0.3032 & 0.928 & 6.6892 & 81.135366M \\ 
        ~ & $\tau$=98[2nd iter.] & ~ & 0.4431 & 0.921 & 5.3029 & 66.959622M \\ 
        ~ & $\tau$=98(freeze)[2nd iter.] & ~ & 0.3947 & 0.925 & 5.129 & 66.959622M \\ \cline{2-7}
        ~ & $\tau$=98[1st iter.] & \multirow{4}{1.5cm}{0.2}  & 0.3383 & 0.933 & 6.7985 & 81.135366M \\ 
        ~ & $\tau$=98(freeze[1st iter.] & ~ & 0.3045 & 0.936 & 6.7756 & 81.135366M \\ 
        ~ & $\tau$=98[2nd iter.] & ~ & 0.5117 & 0.911 & 5.2262 & 66.959622M \\ 
        ~ & $\tau$=98(freeze)[2nd iter.] & ~ & 0.4068 & 0.918 & 5.2201 & 66.959622M \\ \cline{2-7}
        ~ & $\tau$=98[1st iter.] & \multirow{4}{1.5cm}{0.3}  & 0.3808 & 0.928 & 6.8346 & 81.135366M \\ 
        ~ & $\tau$=98(freeze)[1st iter.] & ~ & 0.2683 & 0.942 & 6.824 & 81.135366M \\ 
        ~ & $\tau$=98[2nd iter.] & ~ & 0.3969 & 0.923 & 5.1384 & 66.959622M \\ 
        ~ & $\tau$=98(freeze)[2nd iter.] & ~ & 0.3494 & 0.927 & 5.1097 & 66.959622M \\ \cline{2-7}
        ~ & $\tau$=98[1st iter.] & \multirow{4}{1.5cm}{0.4}  & 0.3836 & 0.921 & 6.7959 & 81.135366M \\ 
        ~ & $\tau$=98(freeze)[1st iter.] & ~ & 0.3071 & 0.931 & 6.737 & 81.135366M \\ 
        ~ & $\tau$=98[2nd iter.] & ~ & 0.4336 & 0.918 & 5.1144 & 66.959622M \\ 
        ~ & $\tau$=98(freeze)[2nd iter.] & ~ & 0.3540 & 0.925 & 5.1172 & 66.959622M \\ \cline{2-7}
        ~ & $\tau$=98[1st iter.] & \multirow{4}{1.5cm}{0.5}  & 0.2765 & 0.937 & 7.0152 & 81.135366M \\ 
        ~ & $\tau$=98(freeze)[1st iter.] & ~ & 0.2304 & 0.94 & 6.9785	 & 81.135366M \\  
        ~ & $\tau$=98[2nd iter.] & ~ & 0.3805 & 0.926 & 5.3181 & 66.959622M \\ 
        ~ & $\tau$=98(freeze)[2nd iter.] & ~& 0.3390 & 0.929 & 5.3052 & 66.959622M \\ \hline
    \end{tabular}}
\caption{Bert - Dair AI Emotion, Comparison with Wanda}\label{suptab:dairaiwanda}}
\end{table*}

\section{Yahoo Answers Topics}

\begin{itemize}
    \item Hareware setting:
    \begin{itemize}
        \item CPU : 14th Gen Intel(R) Core(TM) i9-14900
        \item GPU : NVIDIA GeForce RTX 4090 24GB
        \item MEMORY : 64GB
    \end{itemize}
    \item Dataset : Yahoo answer topics
    \item Model : Bert-base / text classification
\end{itemize}

This experiment compares a model fine-tuned using BERT-base for text classification with 1 epoch to a model compressed by \sysname. Additionally, we conduct experiments with Wanda, comparing the results of applying Wanda alone versus using the integrated tool combining Wanda and \sysname. The sparsity levels for Wanda are 10\%, 20\%, 30\%, 40\%, and 50\%. In this experiment, CKA thresholds of 98 and 99 were used, followed by 3 epochs of retraining and unfreezing/freezing. The batch size was set to 32, the optimizer used was AdamW, and the learning rate was 5e-05. The layer pruning process was based on the similarity of BERT layers, specifically focusing on the encoder. The results of the experiment are as follows.

The results from the basic experiments conducted solely with \sysname\ are shown in Figure~\ref{suptab:yahooat}. Comparison with Wanda is described in Figure~\ref{suptab:yahooatwanda}.

\begin{table*}[t]
{\centering
{\fontsize{8pt}{10pt}\selectfont 
    \begin{tabular}{|l|l|l|l|l|l|l|l|l|}
    \hline
        Stage & CKA Threshold & Num. of enc. & Eval. loss & Accuracy & Eval. time(s) & Training time(s) & Parameter & Freeze \\ \hline
        ~ & ~ & 12 & 1.2287 & 0.7047 & 101.9891 & 41325.0308 & 109.48993 M & ~ \\ \hline
        \multirow{4}{1.5cm}{1st iteration} & $\tau$=99 & 12 (=original) & - & - & - & - & - & - \\ 
        ~ & $\tau$=98 & 9 & 1.094 & 0.710 & 79.4448 & 31593.5162 & 88.226314 M & 0 \\ 
        ~ & $\tau$=99(Part) & 12 (=original) & - & - & - & - & - & - \\ 
        ~ & $\tau$=98(Part) & 9 & 1.1068 & 0.7099 & 80.0802 & 31687.6199 & 88.226314 M & 4 \\ \hline
    \end{tabular}}
\caption{Bert - Yahoo Answers Topics, \sysname\ Only}\label{suptab:yahooat}}
\end{table*}

\begin{table*}[t]
{\centering
{\fontsize{8pt}{10pt}\selectfont 
    \begin{tabular}{|l|l|l|l|l|l|l|}
    \hline
        ~ & CKA threshold & Sparsity level & Eval. loss & Accuracy (w/o training) & Evaluation time(s) & Parameter \\ \hline
        \multirow{5}{1.5cm}{Wanda} & N/A & 0.1 & 1.0265 & 0.71716 & 101.9313 & 109.48993M \\ 
        ~ & N/A & 0.2 & 1.0385 & 0.71616 & 102.1489 & 109.48993M \\ 
        ~ & N/A & 0.3 & 1.03147 & 0.7213 & 102.3629 & 109.48993M \\ 
        ~ & N/A & 0.4 & 1.01104 & 0.7213 & 101.6302 & 109.48993M \\ 
        ~ & N/A & 0.5 & 0.9964 & 0.7214 & 101.568 & 109.48993M \\ \hline
        \multirow{10}{1.5cm}{\sysname,\\[.1ex]Wanda} & Threshold98 & \multirow{2}{1.5cm}{0.1}& 1.0982 & 0.7132 & 79.8796 & 88.226314M \\ 
        ~ & $\tau$=98(freeze) & ~ & 1.0854 & 0.71286 & 79.6928 & 88.226314M \\ \cline{2-7}
        ~ & $\tau$=98 & \multirow{2}{1.5cm}{0.2} & 1.0906 & 0.71483 & 79.9597 & 88.226314M \\
        ~ & $\tau$=98(freeze) & ~ & 1.1071 & 0.7119 & 79.7147 & 88.226314M \\ \cline{2-7}
        ~ & $\tau$=98 & \multirow{2}{1.5cm}{0.3}  & 1.0875 & 0.7126 & 79.6492 & 88.226314M \\ 
        ~ & $\tau$=98(freeze) & ~ & 1.0877 & 0.71203 & 79.8392 & 88.226314M \\ \cline{2-7}
        ~ & $\tau$=98 & \multirow{2}{1.5cm}{0.4} & 1.0789 & 0.71613 & 79.0803 & 88.226314M \\
        ~ & $\tau$=98(freeze) & ~ & 1.0718 & 0.7160 & 79.3074 & 88.226314M \\ \cline{2-7}
        ~ & $\tau$=98 & \multirow{2}{1.5cm}{0.5} & 1.0564 & 0.71763 & 79.4558 & 88.226314M \\
        ~ & $\tau$=98(freeze) & ~ & 1.0523 & 0.7183 & 78.9581 & 88.226314M \\ \hline
    \end{tabular}}
\caption{Bert - Yahoo Answers Topics, Comparison with Wanda}\label{suptab:yahooatwanda}}
\end{table*}

\section{T5 model - SQuAD (QG)}

\begin{itemize}
    \item Hareware setting:
    \begin{itemize}
        \item CPU : intel xeon w5-2455x ( 128 processor )
        \item GPU : NVIDIA RTX A6000 48GB
        \item MEMORY : 256GB
    \end{itemize}
    \item Dataset : SQuAD v1.1
    \item Model : T5-base Question Generation (QG)
\end{itemize}

In this experiment, the SQuAD v1.1 dataset was used with the T5-base model to perform the task of Question Generation. The similarity threshold (CKA threshold) was set to 99\%. After layer pruning, the model was fine-tuned with 3 epochs of retraining. The results of the experiment are in Table~\ref{suptab:t5squad}.

\begin{table*}[t]
{\centering
{\fontsize{8pt}{10pt}\selectfont 
\begin{tabular}{|l|l|l|l|} 
\hline
                 & base-model (T5-base) & 1st\_iter (retrain 3epoch) & final\_iter         \\ 
\hline
Num of encoder   & 12                   & 9                          & 7                   \\ 
\hline
Num of decoder   & 12                   & 5                          & 3                   \\ 
\hline
Num of parameter & 222M (222,903,552)   & 135M (135,588,864)         & 109M (109,630,464)  \\ 
\hline
BLEU             & 0.2268               & 0.2120                     & 0.2009              \\ 
\hline
Precision 1-gram & 0.5319               & 0.5339                     & 0.5296              \\ 
\hline
Precision 2-gram & 0.2809               & 0.2741                     & 0.2659              \\ 
\hline
Precision 3-gram & 0.1825               & 0.1760                     & 0.1678              \\ 
\hline
Precision 4-gram & 0.1237               & 0.1176                     & 0.1103              \\ 
\hline
Brevity Penalty  & 0.9411               & 0.9038                     & 0.8890              \\ 
\hline
Length Ratio     & 0.9411               & 0.9082                     & 0.8890              \\ 
\hline
ROUGE-1 F1       & 0.5177               & 0.5027                     & 0.4929              \\ 
\hline
ROUGE-2 F1       & 0.3033               & 0.2880                     & 0.2765              \\ 
\hline
ROUGE-L F1       & 0.4812               & 0.4697                     & 0.4601              \\ 
\hline
ROUGE-Lsum F1    & 0.4812               & 0.4702                     & 0.4601              \\ 
\hline
METEOR           & 0.4894               & 0.4702                     & 0.4590              \\
\hline
\end{tabular}}
\caption{T5 model - SQuAD (QG).}\label{suptab:t5squad}}
\end{table*}

\end{document}